\documentclass[11pt,a4paper]{article}
\usepackage{xcolor}
\usepackage{soul} 
\makeatletter 
\newcount\SOUL@minus
\makeatother  
\renewcommand\hl[1]{#1} 
\usepackage{graphicx} 
\usepackage{times,latexsym}
\usepackage{url}
\usepackage[T1]{fontenc}
\usepackage{adjustbox}
\usepackage{subcaption}
\usepackage{tablefootnote} 
\usepackage{enumitem} 
\usepackage[acceptedWithA]{tacl2021v1}
\usepackage[]{tacl2021v1}

\usepackage{xspace,mfirstuc,tabulary}

\newif\iftaclinstructions
\taclinstructionsfalse 
\iftaclinstructions
\renewcommand{\confidential}{}
\renewcommand{\anonsubtext}{(No author info supplied here, for consistency with
TACL-submission anonymization requirements)}
\newcommand{\instr}
\fi

\iftaclpubformat 

\else

\fi

\title{DMDD: A Large-Scale Dataset for Dataset Mentions Detection}

\author{
  Huitong Pan$^\diamond$ 
   \ 
  Qi Zhang$^\diamond$ 
  \AND
  Eduard Dragut$^\diamond$ 
  \ 
  Cornelia Caragea$^\dagger$ 
  \ 
  Longin Jan Latecki$^\diamond$ 
  \ \\
  $^\diamond$Temple University, Philadelphia, Pennsylvania, USA
  \\
  $^\dagger$University of Illinois Chicago, Chicago, Illinois, USA
  \\
  \texttt{\{huitong.pan,qi.zhang,edragut,latecki\}@temple.edu}\\
  \texttt{cornelia@uic.edu}
}

\date{}

\begin{document}
\maketitle
\begin{abstract}

The recognition of dataset names is a critical task for automatic information extraction in scientific literature, enabling researchers to understand and identify research opportunities. However, existing corpora for dataset mention detection are limited in size and naming diversity. In this paper, we introduce the Dataset Mentions Detection Dataset (DMDD), the largest publicly available corpus for this task. DMDD consists of the DMDD main corpus, comprising 31,219 scientific articles with over 449,000 dataset mentions weakly annotated in the format of in-text spans, and an evaluation set, which comprises of 450 scientific articles manually annotated for evaluation purposes. We use DMDD to establish baseline performance for dataset mention detection and linking. By analyzing the performance of various models on DMDD, we are able to identify open problems in dataset mention detection. We invite the community to use our dataset as a challenge to develop novel dataset mention detection models.

\end{abstract}

\section{Introduction}

As the volume of scientific literature continues to grow, the automatic extraction of scientific entities from publications becomes increasingly valuable for knowledge management and scientific discovery. In particular, accurate and efficient detection of dataset mentions in the literature can greatly improve the accessibility and usability of scientific data.

Detecting dataset mentions (DMD) presents distinct challenges compared to other Named Entity Recognition (NER) tasks, because the vocabulary used in scientific literature is complex and diverse across different subjects. Domain expertise is often needed to recognize datasets mentioned in the literature. Furthermore, the mention of dataset entities can be potentially ambiguous as the same names may be used to refer to method or task entities. For example, `SGD' is used to represent the stochastic gradient descent method and the Schema-Guided Dialogue dataset. 
Lastly, it is crucial for dataset mentions to include 
linking annotation, which allows linking  Dataset mention to its definition website, such as on GitHub. However, datasets have very diverse ways of being mentioned in the literature, which creates additional challenges for linking. All of these unique issues 
pose non-trivial challenges both to the annotation generation process and to mention recognition models.

There are a number of manually compiled corpora that aim to aid the study of dataset name detection in scientific literature 
\cite{bioNerDS,semeval2017,semeval2018,SCIERC,NLP-TDMS,SciRes,MDER,heddes2021,farber2021identifying}. These corpora have certain limitations. For instance, many of them have only a few hundred instances, which limits their usefulness for training and evaluating dataset mention detection models. Additionally, some of these corpora lack diversity in dataset naming conventions, as they may include mostly capitalized dataset names, whereas many dataset names are not capitalized. Furthermore, these corpora lack entity linking information. Those limitations limit their usefulness for developing dataset mention detection algorithms, despite being suitable for testing such algorithms.

Manual labeling can be prohibitively expensive for NER in general, and for dataset mention detection in particular, because it requires domain expertise. Consequently, there is a need to develop labeled dataset mention data with less human effort to facilitate the training of robust detection models. One potential alternative to manual labeling is to utilize web pages such as GitHub and the Papers with Code (PwC) website\footnote{https://paperswithcode.com/}, which provide metadata about scientific papers. However, these metadata are not always sufficient for building robust dataset mention detection models for two reasons. First, they are often created based on optional manual inputs, which may be prone to human errors and may not be available for new papers or arXiv papers. Second, the metadata is only available at the document level, whereas in-text level mention annotations are required for mention detection models.

In this paper, our objective is to enable the creation of robust dataset detection models by creating a large corpus, called Dataset Mentions Detection Dataset (DMDD). To construct the main corpus of DMDD, we employ distant supervision \cite{mintz} to develop in-paper annotation by taking the dataset mentions from the PwC website and matching them with those in papers. We also add entity linking annotations to each dataset mention. Although the labels obtained by distant supervision may not be as precise as human-generated labels, their vast quantity and diversity of mentions provide an advantage in (pre)training competitive models \cite{emonet, su2019}. DMDD comprises 31,219 scientific articles with over 449,000 dataset mentions automatically annotated in the format of in-text span. Additionally, it includes an evaluation set of 450 scientific papers where we carefully annotated the occurrence of each dataset mention. DMDD corpus can be accessed at the following URL: \url{https://www.kaggle.com/datasets/panhuitong/dmdd-corpus}.
Overall, our paper makes the following contributions: 
\begin{itemize}[noitemsep]
\item We create a large and diverse annotated corpus, the Dataset Mentions Detection Dataset (DMDD), consisting of over 31,000 documents with automatic in-text dataset mention and entity linking annotations.
\item We conduct a comprehensive analysis of existing corpora for dataset mention detection.
\item We establish baseline performance for dataset mention detection and entity linking on DMDD, identify related challenges, and demonstrate the effectiveness of DMDD for training robust models.

\end{itemize}

\section{Related work} \label{sec:related}
\subsection{Related Corpora}
There have been several attempts to create evaluation data for the task of information extraction from scientific papers \cite{farber2021identifying, semeval2017, semeval2018}. 
The followings are the publicly accessible corpora containing annotations for dataset mentions: SciERC~\cite{SCIERC}, SciREX~\cite{scirex}, TDMSci~\cite{tdmsci}, bioNerDS~\cite{bioNerDS}, RCC\footnote{https://github.com/Coleridge-Initiative/rclc} and Heddes~\cite{heddes2021}. These related corpora cover papers from diverse domains, including AI (ML/NLP), biomedical, and social science. 
A few datasets, including bioNerDS and Heddes, were specifically designed for dataset mention detection, whereas others were not. For instance, SciERC and SciREX were also developed for salient entity identification and relation identification tasks, while RCC was initially designed for entity linking.

All of these related corpora are created with manual annotations. In terms of the data generation process, some prior works~\cite{heddes2021, NLP-TDMS} began by sampling instances (e.g., sentences and abstracts) with a high likelihood of containing scientific entity mentions, while others employed trained models~\cite{scirex} or rule-based systems~\cite{bioNerDS} to generate initial noisy annotations before proceeding with manual annotation.
As we show below, DMDD is the first large-scale dataset, and hence more suitable for training SOTA deep learning models.

\begin{figure*}[h]
\includegraphics[width=1\linewidth]{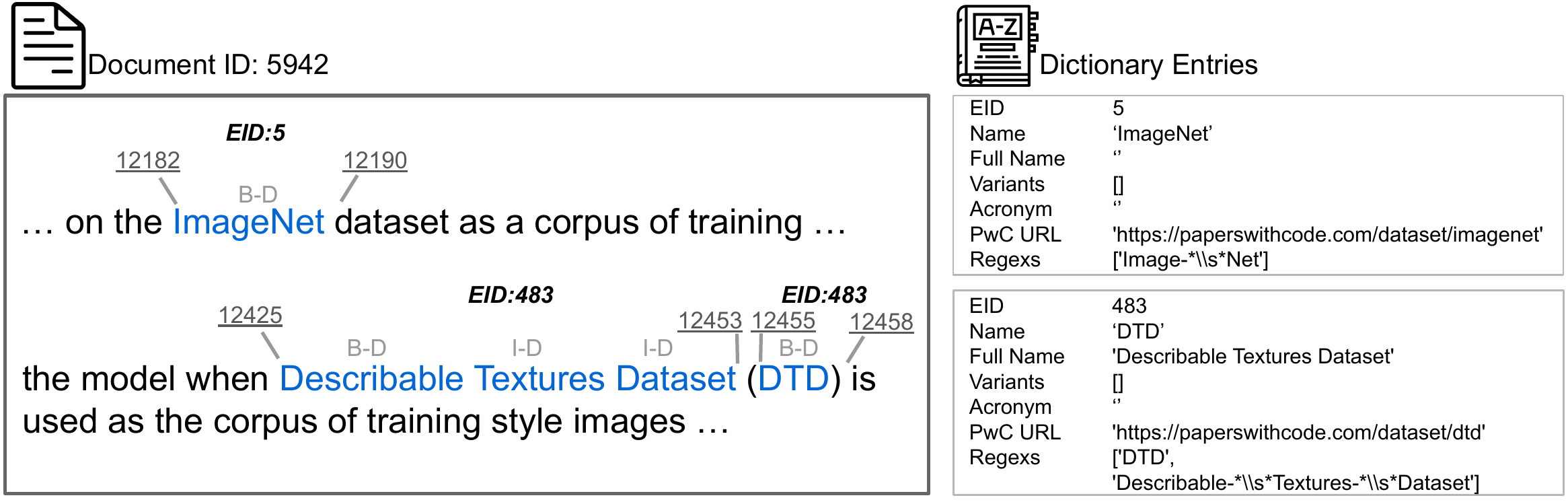}
\caption{\hl{Example of paper-level annotation (left) and dictionary entries (right) in DMDD.} We mark each occurrence of {\color[HTML]{3779D5}dataset (D)} in papers with {\color[HTML]{ababab}\underline{in-text spans} } and \textbf{\textit{entity indexs}}. We can generate the BIO tags. For example, the dataset mention ‘ImageNet’ spans from 12182 to 12190 and has a BIO tag as ‘B-D’.}
\label{fig:data_illustration}
\end{figure*}

\subsection{Dataset Mentions Detection Models}
There already are  approaches for dataset mention detection. For example, \newcite{MDER} proposed utilizing a Long Short Term Memory (LSTM) model and conditional random field (CRF) for this task. Other researchers have examined using a mixture of neural network architectures and pre-trained word embeddings for dataset mention detection. For instance, \newcite{kkim} suggested using Bidirectional LSTM (BiLSTM) with ELMo vectors~\cite{elmo} and GloVe~\cite{glove}. Additionally, \newcite{SciRes} proposed a model that is based on one of the popular choice of NLP models, BERT. Furthermore, \newcite{NLP-TDMS} used a transformer-based method in their TDMS-IE System. 

In addition to the previously mentioned models, there have been several studies on multi-task learning for dataset mention detection. In this approach, the model is trained to jointly perform dataset mention detection and other relevant tasks such as entity linking and relation extraction. For instance, \newcite{scirex} proposed a multi-task model based on BERT that performs scientific mention identification, salient mention detection, pairwise coreference, and salient entity clustering, achieving state-of-the-art performance on the SciREX dataset.

\section{DMDD}
In this section, we provide a detailed description of the development of DMDD's main corpus and DMDD's evaluation set. We also present a comprehensive comparison between DMDD and related corpora.

For the purposes of this study, we define dataset mentions as in-text spans that exclusively comprise the necessary texts (i.e., the name of the dataset) for finding the datasets in knowledge bases. As such, we exclude pronominal references to dataset entities, such as `the dataset' and `the images'. Additionally, it is important to note that the dataset entity in DMDD is distinct from the material entity in other related works, such as SciREX~\cite{scirex}, as the dataset entity is only one of the components of the experimental materials that is annotated as the material entity.

\subsection{DMDD's Main Corpus}
We present the construction of our primary corpus in this subsection. An illustrative example of a DMDD data entry, which includes the parsed scientific article and the in-text annotation for dataset mentions, is displayed in Figure~\ref{fig:data_illustration}.

\subsubsection{Data Collection}
We built DMDD's main corpus by combining data from S2ORC \cite{S2ORC} and Papers with Code (PwC). The parsed scientific articles are obtained from S2ORC \cite{S2ORC}, which is a dataset based on the Semantic Scholar website. S2ORC is a unified resource that combines aspects of citation graphs (i.e., rich paper metadata, abstracts) with a full-text corpus that preserves important scientific paper structure (i.e., sections, inline citation, references to tables and figures). For in-text level annotation of dataset mentions, we used distant supervision to derive the annotations from existing data sources with document-level annotation. 
We sourced the document-level annotation from Papers with Code (PwC), which is a free and open-source website with machine-learning papers, code, datasets, methods, and evaluation tables. For each available paper listed in PwC's data files, we obtained the publication details, PDF web links, and links to related GitHub code. Most of these publication details are edited by the authors of those papers. However, the information about datasets mentioned in the papers is not organized for download. To obtain such information, we conduct web scraping of the `Dataset Section' of each paper's webpage in PwC, which contains human annotations on the document-level about the datasets mentioned in the paper.

\subsubsection{Annotation Procedure}
We describe our distant supervision procedure to create the in-text mention annotation for dataset mentions in this subsection. The document-level annotations 
are based on the data provided by PwC's users. Our premise is that we can take the names supplied by authors in PwC and match them in the main text of a paper. For the most part, this is a correct assumption. However, users do not often give complete information about the artifacts used in their papers. For example, they may only give a partial spelling of an entity name (e.g., `CIFAR' instead of `CIFAR10') or use a different spelling (e.g., `CIFAR-10' in PwC and `CIFAR10' in the paper). Thus, we cannot proceed with a strict matching procedure of dataset names collected from PwC in the text of the papers. 

We commence by creating a dictionary that defines all dataset entities in DMDD. For each dataset entity, we store the following information: name, full name, and web page link in PwC. Next, we create regular expressions (regex) for each dataset entity. The regular expression creation process is described in detail in Section~\ref{sec:regex}. We use regex as an approximate matching procedure to label the parsed text of a paper. Data engineers refer to such data labeling rules as \textit{labeling functions} \cite{NIPS2016_6709e8d6}
. Two example DMDD dictionary entries containing its regex can be found in Figure~\ref{fig:data_illustration}.

Using the document-level annotation on dataset mentions and the regex, we annotated 31,219 scientific articles. For each article, we have the concatenated full-text, section span, document-level dataset annotations, in-text dataset mention span, and the entity index for each mention. Example data can be visualized in Figure~\ref{fig:data_illustration}. 
In addition, we also store section information for each document, which includes the section names and their corresponding starting and ending indices in the concatenated full text. The reason we include section span is that we believe `section' may provide additional semantic information and can impact the detection accuracy. For example, a detection algorithm should be more sensitive to candidates in experiment sections where authors typically describe their datasets.

\subsubsection{Regular Expression Rules}\label{sec:regex}

\hl{
The regex objective is to incorporate as much variety in dataset mentions as possible. However, we do not seek to have an optimal regular expression. First, such a rule is difficult to create manually, and second, we seek to generate enough (weak labeled) data to enable training NER recognizer. Instead of constructing regex for each dataset individually, we use a set of rules to construct regex for all dataset entities, using their short name and full name listed in PwC as base names.

For the 6,675 dataset entities listed in the PwC dataset definition file, there are 8,708 listed name variants. Using an exact match with the base names, we match 7,989 variants. These matched variants are just the short names and full names of the entities. To enhance the exact match, we used a set of rules to customize the regular expression for each base name. The number of additional variants matched with the added rule compared to the exact match is shown as \#Matched.

1) We allow optional space and `-' between words. For example, dataset entity `CIFAR-10' may be mentioned as `CIFAR 10' and `CIFAR10' in papers. To allow such variation, we customize the regex as `CIFAR-*\textbackslash s*10'. (\#Matched = 77).

2) We create acronyms for names including multiple words by combining the initials of the words. For example, we create an acronym `WTQ' for entity `WikiTableQuestions'. (\#Matched = 14).

3) We ignore casing for units appearing in names. In particular, if '3D'/'3k'/'3m' in names, we allow matching '3d'/'3K'/'3M'. For example, dataset entity `DBP15K' may be mentioned as `DBP15k' in papers. To allow such variation,  we customize the regex as `DBP15[Kk]'. (\#Matched = 4).

4) We allow optional decimal places for versions and numbers. For example, the dataset entity `OntoNotes 4.0' may be mentioned as `OntoNotes 4'. To allow such variation, we customize the regex as `OntoNotes 4\textbackslash.*[0-9]*'. (\#Matched = 5).

5) We ignore case for words that have a length greater than 4 and the lowercase of the name is not a common English word. For example, we ignore cases when matching for dataset entity `SciREX', so that it matches `SCIREX' and 'scirex' that may appear in text. We enforce case matching for dataset entity `SHAPES'. (\#Matched = 286). 

6) We allow optional suffixes including `ing' and `ion'. For example, the dataset entity `Deep Soccer Captioning' may be mentioned as `Deep Soccer Caption'. To allow such variation, we customize the regex as `Deep Soccer Captioni*n*g*'. (\#Matched = 0).

7) We allow optional plural forms including `es' and `s'. For example, the dataset entity `MovieLens' may be mentioned as `MovieLen' in papers. To allow such variation, we customize the regex as `MovieLens*'. (\#Matched = 0).

While PwC's listed variants do not include the patterns from rules 6 and 7, we observe many such variations in DMDD's papers caused by typos and loose writing.

 Using all the rules outlined above, we identify names with the corresponding patterns and customized the regex accordingly. This final set of customized regex allows us to cover most of the listed variants, leaving us with only 74 unmatched variants. To address these unmatched variants, we use them as the base names and created additional regex for their corresponding entities.
}

\subsubsection{Data Preprocessing}\label{sec:preprocessing}
With the help of the spaCy python library, we convert the original annotation in the span format to BIO format. 
After the first stage of preprocessing, we discover that we miss some of the annotations for dataset mentions in some sequences. This is because, on PwC websites, the authors or the editors often only annotate the datasets being used in experiments while missing the ones being mentioned. The missing mentions can introduce bias in training as the models may be negatively impacted by learning about the false negative. 
Thus, in the second stage of preprocessing, in order to reduce the number of missing annotations, we combine all regex to search for all possible mentions of the dataset entities in DMDD's dictionary, which was obtained from the PwC website. 
We exclusively apply the second stage of preprocessing to sentences that contain detected dataset mentions from the first stage. This limits the addition of mentions to contexts where the occurrence of dataset mentions is highly likely; this helps mitigate false positives arising from ambiguous entities, such as `SGD'. While `SGD' often appears as a method name in sentences without dataset mentions, it can also appear as a dataset name in co-occurrence with other dataset mentions.

To ensure a consistent comparison between our proposed corpus and existing corpora, we adopted a consistent data preprocessing strategy across all related corpora used in our experiments. In the case of NLP-TDMS and RCC, we used the original text of each paper and their corresponding dataset mention list to develop similar regex patterns to extract dataset mentions in BIO-format. For bioNerDS, the dataset mention span annotations were already provided in BIO-format, so no additional processing was necessary. For SciERC, SciREX, and Heddes, the sequences were already provided in BIO-format annotation.

\subsection{Evaluation Set with Human Annotations}\label{sec:human_annotation}

\begin{table*}[]
\small
\centering
\begin{tabular}{lcccccc}
\hline
\textbf{Corpus} & \textbf{Inst. Unit} & \textbf{\# Inst.} & \textbf{\# Mentions} & \begin{tabular}[c]{@{}c@{}}\textbf{\# Unique} \\ \textbf{Mentions}\end{tabular} & \begin{tabular}[c]{@{}c@{}} \textbf{\# Unique} \\ \textbf{Entities}\end{tabular} & \begin{tabular}[c]{@{}c@{}}\textbf{Entity} \\ \textbf{Linking}\end{tabular} \\ \hline
DMDD (ours) & paper & 31,219 & 449,798 & 10,807 & 6,675 & explicit \\
SciERC~\cite{SCIERC} & abstract & 164 (69) & 770 (122) & 644 (116) & - & - \\
SciREX~\cite{scirex} & paper & 407 & 10,548 & 2,857 & - & - \\
NLP-TDMS~\cite{NLP-TDMS} & paper & 153 & 1,164 & 67 & 99 & explicit \\
TDMSci~\cite{tdmsci} & sentence & 445 & 612 & 478 & - & - \\
bioNerDS~\cite{bioNerDS} & paper & 60 & 920 & 145 & - & - \\
RCC & paper & 2,256 & 36,597 & 1,345 & 1,028 & explicit \\
Heddes~\cite{heddes2021} & sentence & 2,664 & 3,416 & 2,319 & - & - \\ \hline
\end{tabular}
\caption{Summary of corpora for dataset mention detection. The numbers in the brackets for SciERC relate to the corrected version of SciERC without annotation errors.}
\label{tab:summary_corpora}

\end{table*}

We manually annotated two sets of instances for evaluation purposes, one set is from DMDD and the other is from SciREX. As SciREX provides publicly available manually annotated documents with scientific entities, we only needed to refine their annotations to meet DMDD's standards. All evaluation sets were manually annotated by three NLP researchers using brat rapid annotation tool~\cite{brat}. We aggregated the annotations by keeping the mentions where at least two annotators agreed.

For the DMDD evaluation set (DMDD-E), annotators were tasked with manually annotating 450 papers that were sampled from DMDD's test set. The annotators were instructed to verify the detected mentions from DMDD's main corpus and identify any missing mentions in each paper. Additionally, they were required to verify the entity linked to each mention. To ensure accuracy, annotators were directed to search the PwC website and Google to confirm dataset entities during the annotation process.

To assess the level of agreement between annotators, we used the relaxed span matches method, which considers a match when the dataset mention spans from the three annotators overlap. The resulting Fleiss kappa of 0.79 represents a substantial agreement between annotators. DMDD's evaluation set contains 13,039 mentions for 682 DMDD entities, with 1,964 mentions that could not be linked to the DMDD dictionary. On average, each annotator required approximately 15 minutes to annotate a pre-annotated paper with weak labels.

When compared to DMDD's evaluation set, the weak labels from DMDD's main set obtains an F1 score of 77.9\%, recall of 68.1\%, and precision of 91.2\%. The low recall indicates that most of the weak labeling errors are due to missing dataset mentions. We identify two main reasons for the missing mentions. First, mentions may contain rarely-used version names that distant supervision provides only partial annotation for, such as `KITTI 2012', where only `KITTI' is tagged and the version part of the name, i.e., `2012', is ignored. Second, missing mentions may occur in contexts without mentions of the document-level annotated dataset, such as in related work sections where only one dataset is mentioned, or in sentences where the dataset is mentioned by itself as a pre-trained dataset in the description of methods.

\subsection{Comparison with Related Corpora}
We compare DMDD with seven related corpora containing dataset mentions annotations in Table\ref{tab:summary_corpora}, where `Inst.' is used to represent `instance' and `\#' is used to represent `number'. In order to compare corpora fairly, we exclude the negative instances from the calculation of `\# Instances', as some corpora do not contain negative instances.

\subsubsection{Corpora Size} 
DMDD has the largest size among the discussed corpora, in terms of the number of instances (\# Inst. = 31K), instance unit (Inst. Unit = Paper), and the number of mentions (\# Mentions = 450K). With paper-level annotations, DMDD allows for a larger input unit, such as a section, which can provide richer context and potentially benefit mention detection models. 

SciERC samples instances from abstracts. Sampling instances from a specific section of papers may create corpora with limited variation in lexical and syntactic expressions (for example, the language of abstract sections is different from that of methodology sections). 
A benefit of DMDD over most of the other existing corpora is that an entity mention appears in multiple sentences across the 31K papers, offering diverse context learning opportunities in training. This is captured by the number of unique mentions and the number of mentions in Table~\ref{tab:summary_corpora}. 
While the related corpora give better-labeled data (because they're manually created), their data annotation processes are not scalable since they heavily depend on manual labeling.

\subsubsection{Diversity of Dataset Mentions} \label{sec:diversity}

Intuitively, dataset names (e.g., `CIFAR10') that consist of a single word, that include capitalized letters, and do not have non-literals are easy to detect. However, many dataset names do not follow this pattern. They may contain non-literals (e.g., `YUP++'),  may not be capitalized (e.g., `iris'), or may contain multiple words (e.g., `Atomic Visual Actions'). Such diversity of dataset naming poses detection difficulties. A (training) corpus needs to avoid being biased toward any of such categories and contain enough samples from each category. We perform an in-depth analysis of all annotated dataset mentions in related corpora to examine the diversity of mentions. 

\begin{table}[]
\centering
\begin{adjustbox}{width=1\linewidth}
\small
\begin{tabular}{rcccccc}
\hline
\textbf{Corpus} & \multicolumn{2}{c}{\textbf{\begin{tabular}[c]{@{}c@{}}Long\\ Mention\end{tabular}}} & \multicolumn{2}{c}{\textbf{\begin{tabular}[c]{@{}c@{}}Alpha. \& \\ Punct. \end{tabular}}} & \multicolumn{2}{c}{\textbf{\begin{tabular}[c]{@{}c@{}}All\\ Lower\end{tabular}}} \\
 & \# & \% & \# & \% & \# & \% \\
 \hline
DMDD & 3,044 & 28 & 7,903 & 73 & 1072 & 10 \\
SciERC & 552 & 86 & 612 & 95 & 353 & 55 \\
SciERC$_{C}$ & 7 & 10 & 50 & 72 & 1 & 1 \\
SciREX & 2,122 & 74 & 2,102 & 73 & 307 & 11 \\
NLP-TDMS & 48 & 48 & 60 & 61 & 0 & 0 \\
TDMSci & 335 & 70 & 317 & 66 & 10 & 2 \\
bioNerDS & 34 & 31 & 104 & 95 & 3 & 3 \\
RCC & 2,869 & 91 & 2,469 & 78 & 71 & 2 \\
Heddes & 2,161 & 83 & 1,774 & 68 & 81 & 3 \\
\hline
\end{tabular}
\end{adjustbox}
\caption{Distribution of different types of dataset mentions in DMDD and existing corpora. \# and \% indicate the number and percentage of the corpus' unique mentions exhibiting certain characteristics. SciERC$_{C}$ represents the corrected version of SciERC without annotation errors.}
\label{corpora_features}
\end{table}

For each corpus, we perform the following evaluation steps and summarize the evaluation results in Table~\ref{corpora_features}. First, we extract all in-text mentions of the dataset names, using the provided annotations. We derive the unique mentions from all the in-text mentions. Notably, unique mentions do not equal unique datasets as one dataset may be referred to as different text strings (e.g., `MHP' may be referred to as `Multiple-Human Parsing'). Second, we find mentions with different characteristics, which are defined as follows.

1) Long mentions. If the mention contains white spaces, then it is a long mention containing multiple words. 
This is important as long mentions are often harder to be detected accurately than single-word mentions. 

2) Character level composition. 
Alphabet and Punctuation Only (Alpha. \& Punct.): check if the mention contains only alphabet and punctuation. We want to see the number of mentions containing no numerical characters. From a reader standpoint, it is often easier to classify entities with a combination of alphabets and numerical values (e.g.: `MediaEval2010') as dataset names than those without (e.g.: `English-Hungarian'). 

3) Capitalization. 
All Lower-cased (All Lower): We seek to account for dataset names with all the characters being lowered-cased in a dataset mention. As commonly agreed, words including upper case letters often indicate that they are specialized words and are more likely to be dataset mentions than those without upper case letters.

As shown in Table~\ref{corpora_features}, with the exception of DMDD, NLP-TDMS, and bioNerDS, the available corpora demonstrate an imbalanced distribution that skews towards long mentions. SciERC and bioNerDS, in particular, exhibit a prevalence of mentions that consist solely of letters and punctuation, with only a small fraction containing numeric characters. Additionally, with the exception of SciERC, all corpora are inclined towards mentions that feature uppercase letters. Hence, individually none of them have enough unique mentions from each category to enable training of robust models across all categories. We also note that the characteristics presented in Table~\ref{corpora_features} are non-exhaustive, non-exclusive, and may overlap.

\subsubsection{Entity Linking} \label{sec:linking_capability}

Entity Linking (EL) for datasets is the task of associating a dataset mention in text with a dataset entity in a knowledge base, such as Papers with Code. The entity linking information for dataset mention is important as it enables users to refer to the right dataset or download the correct dataset for empirical studies. We distinguish two categories of linking: explicit linking and non-linking. We categorize the type of linking for existing corpora in Table~\ref{tab:summary_corpora}. We note that in Table~\ref{tab:summary_corpora}, the "-" symbol represents non-linking. 

DMDD is created based on PwC and each entity mentioned in DMDD’s main corpus has an explicit link to the PwC website with a unique identifier. RCC and NLP-TDMS also have explicit linking since they provide URL links to the knowledge bases with dataset information. Specifically, all the datasets from RCC can be linked to ICPSR\footnote{https://www.icpsr.umich.edu/web/pages/} and all the datasets in NLP-TDMS can be linked to NLP-Progress\footnote{https://nlpprogress.com/}. However, all the other corpora do not provide such explicit linking information.

\begin{table}[]
\small
\begin{tabular}{r|l}
\hline
\textbf{Corpus}        & \textbf{Examples}                   \\ \hline
DMDD                   & `MNIST', `General Language          \\
                       & Understanding Evaluation \\
                       & benchmark' \\ \hline
SciERC                 & `image data', `written texts',      \\ \hline
SciREX                 & `SQuAD) ',                          \\
                       & `augmented PASCAL train set'        \\ \hline
NLP-TDMS               & `SemEval-2010 Task 8',              \\
                       & `Quora Question Pairs'              \\ \hline
TDMSci                 & `forums', `a separate set           \\
                       & of 40 ACE 2005 newswire texts'      \\ \hline
bioNerDS               & `String', `Gene Ontology'           \\ \hline
RCC                    & `balance sheet data',               \\
                       & `External Position Report'          \\ \hline
Heddes                 & `MNIST or the ImageNet dataset',    \\
                       & `text datasets'                     \\ \hline
\end{tabular}
\caption{Dataset mention annotation examples from DMDD and existing corpora.}
\label{tab:ann_example}
\end{table}

For the related corpora without explicit linking information, we attempted to link their annotated mentions to PwC and the other websites, like the ACL Anthology, but we were unsuccessful in linking a significant portion of the annotated mention. In addition, our early empirical studies with these corpora showed an unexpectedly low recall rate on detecting dataset mentions, which prompted us to manually verify some of the data. We asked two Ph.D. students with NLP expertise to manually go over the annotated data in SciERC. It was not our goal to verify all data sources, which would have taken substantial labor. Table~\ref{tab:ann_example} shows some example dataset mention annotations for related corpora. We identify four potential reasons contributing to the failure of linking. We exemplify them using mentions from SciERC. 

1) Mentions include extra characters or text strings [9 (1\%)]. For example, the mention `aligned wordnets' includes the descriptive text `aligned' for the datasets. Additionally, in the original document, this mention actually refers to multiple wordnets that are being aligned by the proposed method. In Table~\ref{tab:ann_example}, `SQuAD)' includes the extra character `)' which may be the result of human error.

2) Mentions include more than one dataset [8 (1\%)]. For example, `SemCor and Senseval-3 datasets'. 

3) Mentions do not include the actual dataset name [559 (87\% )], for example, `records' and `CD-covers'. This is because some related corpora are annotated with pronominal reference to entities, as defined in ACE 2005~\cite{ace2005}. Pronominal reference is not helpful in linking mentions to dataset entities, especially when the corpora are not annotated on the paper level and the proper name reference is missing from the annotated instance. Within this characteristic group, there are also confusing mentions not using the most commonly-used dataset names or missing part of the names [5 (1\%)]. For example, `treebank' can denote many possible datasets, such as The Penn Treebank~\cite{10.5555/972470.972475} and CHILDES Treebank~\cite{Pearl2013SyntacticIA}. This further points toward the need to include linking attributes in the annotation whenever possible.

Among all of the unique mentions in SciERC, only 69 (11\%) do not exhibit the three discussed characteristics. As shown in Table \ref{tab:summary_corpora}, when only considering the correct mentions, the number of mentions and instances with mentions are significantly reduced. Also, as shown in Table \ref{corpora_features} for SciERC$_{C}$, the percentage of long mentions and all-lower-case mentions drops significantly, yielding a more biased set of dataset mentions.

All existing corpora, except NLP-TDMS, share similar characteristics to SciERC. NLP-TDMS follows the NLP-Progress taxonomy website to annotate their entities, which means all the dataset names they used for labeling their instances are actual dataset names. 

In contrast to the existing corpora, DMDD has the following advantages. DMDD is the largest corpora with more than 31K instances. DMDD has the largest number of mentions and the largest number of unique mentions, providing more mention examples than existing corpora. In terms of the diversity of dataset mentions, DMDD exhibits some biases on having a small percentage of all-lower cased mentions. However, since DMDD contains a significantly larger amount of mentions and unique mentions than existing corpora, DMDD can still provide enough examples with different characteristics. In terms of entity linking, all DMDD's annotated mentions can be directly linked to Papers with Code web pages.

\section{Experimental Setup}
The experiments are designed to address the task of dataset entity mentions and entity linking, with three primary objectives in mind: establishing baseline performance on our dataset, providing insights into the difficulty of each task, and evaluating the effectiveness of using DMDD for training.

\subsection{Mention Detection}~\label{sec:Mention Detection description}
We formulate the task of dataset mention detection as a token-level tagging task, and evaluate a broad range of models as baselines in our experiments. To explore the impact of input size, we evaluate models with different input lengths. Since most existing approaches for dataset mention detection operate at the sentence-level, we split the models into two categories: sentence-level models and beyond sentence-level models. 

\subsubsection{Sentence-Level}
We conducted experiments on sentence-level inputs using various models, including Conditional Random Fields (CRF), Bidirectional Long Short-Term Memory (BiLSTM), BERT~\cite{bert_base_cased}, and SciBERT~\cite{beltagy-etal-2019-scibert}. For the CRF model, we used features that incorporate Part-of-Speech (POS) tags and keywords~\cite{heddes2021}.

For BERT and SciBERT, we used the pretrained weights: base-cased BERT~\cite{bert_base_cased} and scivocab-cased SciBERT~\cite{beltagy-etal-2019-scibert}. Then, we fine-tuned them on our training corpora. All hyperparameters used for training the models were the same as in the original SciBERT~\cite{beltagy-etal-2019-scibert}, except for the batch size, which was set to 16.

For BiLSTM, we evaluated two additional variations: BiLSTM-G and BiLSTM-W, which utilize pre-trained embeddings initialized with GLoVe~\cite{glove} and Word2Vec~\cite{word2vec}, respectively. We loaded both pre-trained embeddings using the Gensim Python library and initialized tokens that were not mapped with pre-trained embeddings to zeros. The embedding layer was updated during training for all tokens. To ensure a fair comparison, we used a 300-dimensional embedding layer for BiLSTM, BiLSTM-G, and BiLSTM-W.

For BiLSTM-G, we used the embedding trained on Wikipedia and Gigaword, converting 30,428 tokens in the entire corpus, while 120,190 tokens were missing from the pre-trained embeddings. We observed that most dataset names were missing from the pre-trained embeddings.

Similarly, for BiLSTM-W, we used the embedding trained on Google News, converting 63,321 tokens while 87,297 were missing. We hypothesize that by incorporating additional learned semantic information from large corpora, these two versions of BiLSTM can outperform the regular BiLSTM in predicting dataset name mentions.

\subsubsection{Beyond Sentence-Level}
\hl{
To evaluate model input sizes beyond sentence-level, we examined two models optimized for longer sequence length: SciBERT and LongFormer~\mbox{\cite{beltagy_peters_cohan_2020}}. Additionally, we evaluated two different input sizes, section-level and 512-tokens-level. For the section-level inputs, we cropped the documents based on their sections, whereas for 512-tokens-level inputs, we cropped the documents to sequences with a fixed length of 512 tokens. Notably, some of these sequences contain dataset mentions while others do not.
}
\subsection{Entity Linking}

\hl{
Entity linking (EL) for dataset entities, as a special subproblem of EL, differs from the typical general EL task, which links general entities into a huge knowledge base (KB) like Wikipedia. In our work, we utilize PwC as the KB, which contains 7,795 entities. To evaluate the EL task on our dataset, we conduct baseline experiments for EL using two methods. 
Specifically, we consider the EL given true spans, then we take the span of the dataset mention as the query, and PwC as the KB. We then utilize an information retrieval approach to retrieve the top $K$ most relevant dataset entities in the KB. We conduct experiments in both sparse retrieval and dense retrieval using Pyserini \cite{lin2021pyserini}. In Pyserini, sparse retrieval is based on BM25 and uses bag of word representations, while dense retrieval employs transformer-encoded representations, with the encoder being ColBERTv2 \cite{santhanam2021colbertv2}. All parameters use the default settings of Pyserini.}

\subsection{Train-Test Split} 
For DMDD and all the corpora used in our experiments, we first perform a train-test split at the document level. Subsequently, we perform a train-test split at other levels, such as section-level and sentence-level, based on the document-level split. For DMDD, we used 70\% of the documents for training and 30\% for testing. 

The DMDD-E set, which is a manually annotated test set of 450 documents, was sampled from the DMDD's test set. We report results on this set in our paper. The DMDD-E set contains a zero-shot subset consisting of 10 dataset entities. These zero-shot entities were randomly selected from DMDD, and none of them appear in any corpus's training set.

When training mention detection models, we use a split of 80\% positive sequences and 20\% negative sequences in most of the experiments, unless otherwise specified. The goal of negative sentences is to balance the fact that we only consider one type of NER and to facilitate better generalization for deep learning models. In particular, we seek to avoid false positive predictions, since the majority of sentences in scientific papers contain no dataset mentions. Table~\ref{tab:nSequences} summarizes the median sequence length in tokens and the number of sequences containing dataset mentions in DMDD.

\begin{table}[]
\begin{adjustbox}{width=1\linewidth}
\begin{tabular}{lrrrr}
\hline
\multicolumn{1}{l|}{} & \begin{tabular}[c]{@{}r@{}}\textbf{Median} \\ \textbf{Length}\end{tabular} & \textbf{N. All} & \textbf{N. Train} & \textbf{N. Test} \\ \hline
\multicolumn{1}{l|}{Sentence} & 30 & 792,554 & 532,349 & 260,205 \\
\multicolumn{1}{l|}{Section} & 372 & 245,506 & 167,954 & 77,552 \\
\multicolumn{1}{l|}{512-Token} & 512 & 150,207 & 101,969 & 48,238 \\ 
\multicolumn{1}{l|}{Document} & 5,729 & 31,210 & 21,847 & 9,363 \\
\hline
\end{tabular}
\end{adjustbox}
\caption{The median sequence length in tokens and the number of sequences containing dataset mentions in DMDD.} 
\label{tab:nSequences}
\end{table}

\begin{table*}[h!]
\centering
\begin{adjustbox}{width=1\textwidth}
\begin{tabular}{r|ccc|ccc|ccc}
\hline
\multicolumn{1}{l|}{} & \multicolumn{3}{c|}{Positive and Negative} & \multicolumn{3}{c|}{Positive} & \multicolumn{3}{c}{Zero-Shot} \\ \hline
\# Sentences & \multicolumn{3}{c|}{10,722} & \multicolumn{3}{c|}{8,602} & \multicolumn{3}{c}{89} \\ \hline
Model & F1 & Precision & Recall & F1 & Precision & Recall & F1 & Precision & Recall \\ \hline
CRF & .681 ± .000 & .550 ± .000 & .893 ± .000 & .682 ± .000 & .550 ± .000 & .898 ± .000 & .342 ± .000 & .215 ± .000 & .842 ± .000 \\
BiLSTM & .647 ± .013 & .546 ± .020 & .795 ± .009 & .650 ± .014 & .546 ± .020 & .802 ± .006 & .256 ± .020 & .168 ± .012 & .550 ± .096 \\
BiLSTM-G & .652 ± .012 & .548 ± .017 & .804 ± .004 & .653 ± .012 & .548 ± .017 & .810 ± .004 & .268 ± .041 & .181 ± .036 & .522 ± .061 \\
BiLSTM-W & .594 ± .019 & .498 ± .019 & .739 ± .017 & .596 ± .017 & .498 ± .019 & .746 ± .003 & .258 ± .037 & .175 ± .024 & .511 ± .041 \\
BERT & \textbf{.751 ± .006} & .635 ± .009 & \textbf{.920 ± .002} & .753 ± .006 & .635 ± .009 & \textbf{.926 ± .002} & .572 ± .012 & .417 ± .012 & \textbf{.907 ± .004} \\
SciBERT & \textbf{.751 ± .002} & \textbf{.639 ± .002} & .912 ± .002 & \textbf{.754 ± .002} & \textbf{.639 ± .002} & .919 ± .002 & \textbf{.586 ± .008} & \textbf{.436 ± .011} & .898 ± .010 \\ \hline
\end{tabular}
\end{adjustbox}
\caption{The performance of mention detection models with sentence-level input.}
\label{tab:SOTA}
\end{table*}

\section{Experimental Results}
This section describes the experimental setup and results of the conducted experiments.

\subsection{Mention Detection}\label{sec:SOTA-results}
All mention detection models discussed in Section~\ref{sec:Mention Detection description} have been trained in 3 rounds with randomly shuffled training sets of DMDD. The average and standard deviation of scores are calculated based on the exact match.

\subsubsection{Sentence-Level Performance}
We evaluate the performance of the mention detection models with sentence-level inputs on three sets: the full set of DMDD-E, the positive subset of DMDD-E, and the zero-shot subset. DMDD-E's full set comprises 80\% positive and 20\% negative sequences, with positive sequences including all occurrences in the documents and negative sequences randomly drawn from the documents. Including the negative sequences allowed us to assess the models' ability to accurately classify both positive and negative sentences, which is crucial for real-world applications where the presence of a dataset mention may be rare. Model performance scores, including F\textsubscript{1} score, precision, and recall, were computed based on exact match and are shown in Table \ref{tab:SOTA}. It is important to note that the relative importance of precision and recall may vary depending on the specific use case and application. For example, precision may be more important in scenarios where false positives can have significant consequences, as it may reduce the reliability of the tool and potentially lead to erroneous analysis or decision-making. On the other hand, in scenarios where missing a dataset mention may lead to missed opportunities for data analysis, recall may be more important.

Overall, SciBERT and BERT performances are close. They have the top performance across all the evaluation metrics in all evaluation sets. 

One interesting finding is that the CRF model outperforms the BiLSTM models in our experiments. This can be attributed to the CRF implementation \cite{heddes2021}, which incorporates expert-designed features that leverage part-of-speech tags and capitalization patterns; they  are particularly informative in detecting dataset mentions in scientific literature. In contrast, BiLSTM models rely entirely on learned features, which may not be as effective in capturing the unique nuances of dataset entities.
For BiLSTM, the model variation using Word2Vec embedding (BiLSTM-W) and the model variation using GloVe embedding (GloVe) perform similarly to the original version of BiLSTM and bring no significant performance improvement. 

\subsubsection{Beyond Sentence-Level Performance}
\begin{table}[]
\begin{adjustbox}{width=1\linewidth}
\begin{tabular}{rccc}
\hline
\multicolumn{1}{r|}{Model} & F1 & Precision & Recall \\ \hline
\multicolumn{4}{c}{Sentence-Level Input} \\ \hline
\multicolumn{1}{r|}{SciBERT} & .016 ± .003	&.302 ± .059 & .008 ± .002 \\\hline
\multicolumn{4}{c}{Section-Level Input} \\ \hline
\multicolumn{1}{r|}{Longformer} & .731 ± .004 & .625 ± .005 & .881 ± .002 \\
\multicolumn{1}{r|}{SciBERT} & \textbf{.732 ± .003} & .619 ± .003 & \textbf{.897 ± .000} \\ \hline
\multicolumn{4}{c}{512-Token-Level Input} \\ \hline
\multicolumn{1}{r|}{Longformer} & .695 ± .004 & \textbf{.661 ± .006} & .733 ± .005 \\
\multicolumn{1}{r|}{SciBERT} & .698 ± .002 & .652 ± .006 & .750 ± .009 \\ \hline
\end{tabular}
\end{adjustbox}
\caption{\hl{The performance of mention detection models with different input sizes when evaluating on full documents.}}
\label{tab:beyond}
\end{table}

\hl{
For models beyond sentence-level, we crop each evaluated document into overlapped sequences. Specifically, we used a 5\% overlap between adjacent sequences. Then, we mapped the predicted results for each sequence back to document-level for evaluation purposes. We used argmax when computing the predicted results for the overlapping tokens. Table~\mbox{\ref{tab:beyond}} presents the performance of mention detection models with sentence-level, section-level input, and 512-token-level input on DMDD-E. The table showcases the F1 score, precision, and recall metrics, which are computed based on exact match. 

The evaluation of the sentence-level model on entire documents in Table~\mbox{\ref{tab:beyond}} shows significantly lower performance than the evaluation on mostly positive sentences containing mentions in Table~\mbox{\ref{tab:SOTA}}. This highlights the challenges of sentence-level models in dealing with the highly sparse dataset mentions in scientific literature.

When considering input sizes beyond the sentence-level, we observed that SciBERT performed comparably to LongFormer.
Furthermore, models trained with section-level input have superior performance compared to those trained with 512-token-level input. This may be attributed to the higher density of dataset mentions in section-level input, as sections are generally shorter than 512 tokens. This finding is also supported by the data presented in Table~\mbox{\ref{tab:nSequences}}.
The improved performance of section-level models may also suggest that splitting based on sections provides additional semantics that is advantageous for training when compared to splitting based on 512-token lengths, which ignores the semantic structure of the documents.
}
\begin{figure*}[h!]
\includegraphics[width=1\linewidth]{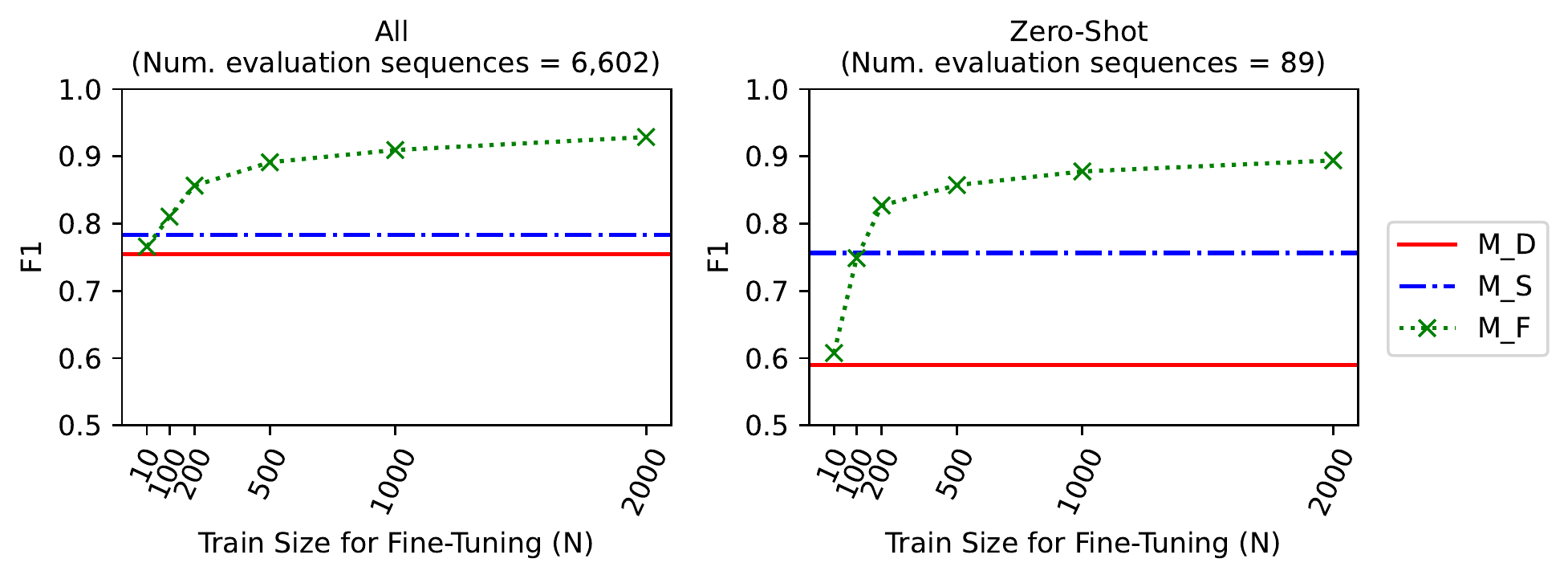}
\caption{\hl{Trend of F1 when varying the number of human annotations.}}
\label{fig:finetune}
\end{figure*}

\subsubsection{Error Analysis}~\label{sec:error_analysis}
\begin{table}[t]
\centering
\small
\begin{tabular}{r|cccc}
\hline
Category & N & F1 & P & R \\ \hline
Long Sequences & 1,808 & 0.66 & 0.54 & 0.85 \\
Multiple mentions & 4,326 & 0.69 & 0.55 & 0.91 \\
Unseen entities & 1,650 & 0.54 & 0.39 & 0.84 \\ \hline
\end{tabular}
\caption{SciBERT model performance on subsets of DMDD-E with instances in different categories. N represents the number of tested sequences in the related category.}
\label{tab:results-category}
\end{table}
Based on the performance of sentence-level inputs, we conduct an error analysis on the SciBERT model and aim to identify common patterns among the erroneous instances. As shown in Table~\ref{tab:SOTA}, we observe that consistently the models have low precision and high recall, indicating a high number of false positives. After analyzing the false positives, we find that the model frequently misclassified mentions such as `SGD' that have ambiguous meanings.

In addition, we identify three common patterns: long sequence length, multiple mentions, and unseen entities. The category of unseen entities includes not only the 10 zero-shot entities but also entities that are labeled by human annotators but cannot be linked to the DMDD dictionary. None of the unseen entities has any annotated mention in the training dataset. 

Table \ref{tab:results-category} presents the F1, precision (P), and recall (R) of the SciBERT model on subsets of DMDD-E, grouped by the common patterns identified earlier. Performance scores are computed based on the exact match. SciBERT performed worse than average on all common patterns, with the poorest performance in the unseen category. This is consistent with the zero-shot performance presented in Table~\ref{tab:SOTA}.

\subsubsection{Fine-Tuning with Strong Labels}

To evaluate the efficacy of DMDD for training purposes, we conduct a comparative analysis of SciBERT models that are trained solely with weak labels from DMDD and those trained solely with human labels from SciREX.  We also examine the minimum number of human labels required to fine-tune a model for achieving a similar level of performance. We split the DMDD-E and SciREX into training sets (DMDD-E-Tr and SciREX-Tr) and testing subsets (DMDD-E-Te and SciREX-Te), where all sequences containing zero-shot entities are allocated to the testing set. We do not train or test with negative sequences, which contain no dataset mention. This is done to investigate the effect of fine-tuning using human labels while isolating the influence from negative samples. 

We developed three types of SciBERT models, as follows:
(1) M\_D, which is trained using DMDD;
(2) M\_S, which is trained using SciREX-Tr, which has 4900 manual annotated sequences;
(3) M\_F, which is fine-tuned on top of M\_D using N sequences that are randomly sampled from DMDD-E-Tr. We conduct experiments with different N values, including 10, 100, 200, 500, 1000, and 2000.

All models are then evaluated on DMDD-E-Te. Figure~\mbox{\ref{fig:finetune}} depicts the performance of the models and the F1 trend when varying the number of human annotations. The performance patterns from the overall testing set and the zero-shot subset are similar. As anticipated, the model (M\_D) trained with only weak labels underperforms the models (M\_S) trained with human-annotated labels. We observed that for M\_F, fine-tuning with 100 strong labels enables a better performance than M\_S, which is trained solely with strong labels. In other words, fine-tuning the pre-trained model from DMDD with approximately 5 human-annotated documents yields a performance similar to the model trained with around 245 human-annotated documents. Furthermore, fine-tuning with 1,000 human-annotated sequences leads to a further improvement in performance, achieving 0.9 F1 scores on DMDD-E-Te.

\subsubsection{Train Size vs. Performance} \label{sec:exp-results-train-size}
\begin{figure}[]
\centering
\includegraphics[width=0.8\linewidth]{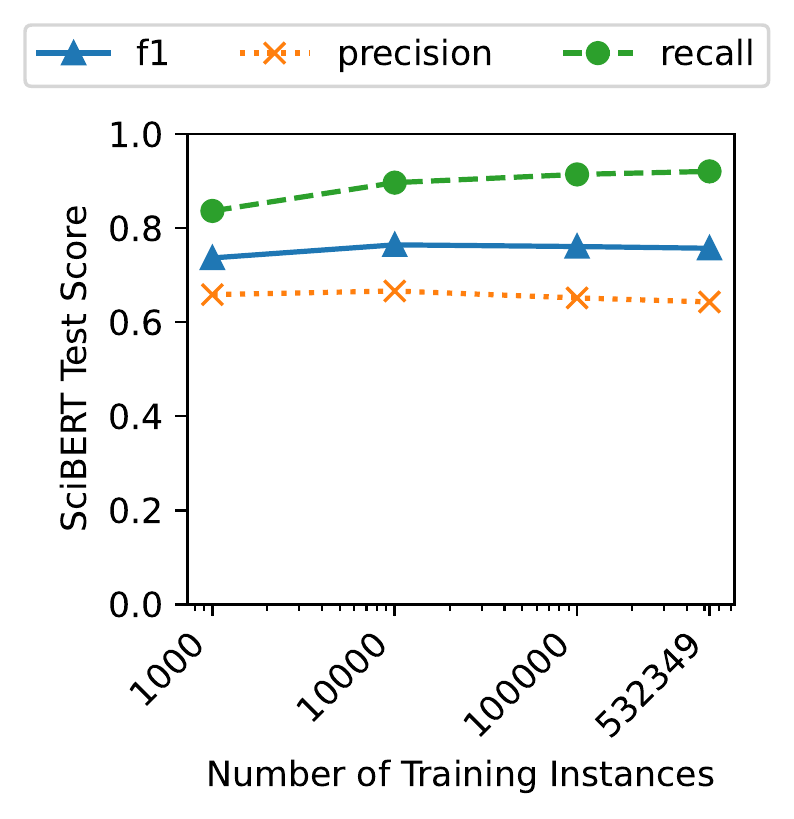}
\caption{Test performance of SciBERT when training on DMDD as the train size increases.}
\label{fig:trainsize}
\end{figure}

As part of our ablation study, we investigate the training benefits resulting from the large size of DMDD. To this end, we trained SciBERT on sentence level using different sizes of DMDD, while maintaining an 80\%-20\% ratio between positive sequences and negative sequences. We then evaluate the trained models using DMDD-E and calculate their performance scores based on exact match. The results are presented in Figure~\ref{fig:trainsize}.

Our analysis reveals that the most significant improvement in model performance occurs when increasing the training size from 1000 to 10000. The recall score continues to improve as the training size increases, while the F1 score and precision remain stable beyond a training size of 10000. This suggests that the model is predicting more false positives when the training size increases. To better leverage the large size and the diverse mentions in DMDD and enhance the model's performance, it can be beneficial to balance the training datasets before training. For instance, sampling more samples with the common features of the challenging cases discussed in Section~\ref{sec:error_analysis} can be a fruitful strategy.

\subsection{Entity Linking} 

\begin{table}[]
\begin{adjustbox}{width=1\linewidth}
\centering
\small
\begin{tabular}{l|lllll}
\hline
EL method & R@1    & R@3    & R@5    & R@10   & R@50   \\ \hline
BM25      & 0.340 & 0.531 & 0.541 & 0.541 & 0.720 \\
ColBERTv2   & 0.354 & 0.550 & 0.578 & 0.632 & 0.726 \\ \hline
\end{tabular}%
\end{adjustbox}
\caption{\hl{Entity linking performance evaluated by recall with top $K$ entity (R@$K$).}} 
\label{tab:EL_Results}
\end{table}

\hl{
Table \mbox{\ref{tab:EL_Results}} presents the experimental results for the Entity Linking (EL) task on our dataset, employing both sparse retrieval (BM25) and dense retrieval (ColBERTv2) methods. 
Despite not being fine-tuned, ColBERTv2 outperforms BM25, particularly in terms of R@10. However, there remains significant potential for model improvement in EL for dataset entities.
For BM25, most of the errors occur due to the mentioned abbreviations that never appear in the KB. For instance, researchers may use `H3.6M' to represent the `Human3.6m' dataset, but this abbreviation never appears in any entity's description text in the KB. 
For ColBERTv2, many errors occur when the sentences with dataset mention are not descriptive of the dataset, making it difficult for the model to disambiguate based on context. An example is the sentence ‘We test our method on H3.6M’.
}

%


\section{Limitations and Future Work}
The DMDD corpus is annotated through distant supervision, which prioritizes scale over accuracy. The current scope of DMDD is limited to dataset mentions that can be linked to the DMDD dictionary, resulting in missing labels for dataset mentions that are not listed on PwC websites or that have variations not included in the regular expression. This limitation may introduce annotation noise, especially when dealing with dataset subversions that are not explicitly listed in PwC. Furthermore, DMDD does not include annotations for ambiguous cases, where distinct datasets have the same name or share acronyms, nor does it consider changes in naming conventions over time. Similar limitations apply to other corpora created using distant supervision, as annotation accuracy heavily relies on manual correction. To address these limitations, future work can focus on developing more advanced methods for mention detection and exploring alternative approaches to distant supervision. Additionally, DMDD can be extended to include annotations for more challenging test instances, such as unseen mentions, ambiguous mentions, and mentions with diverse sub-versions. We also plan to periodically release revised versions that have larger sizes and additional (manual) annotations for scientific entities such as model and method names.


In terms of model performance, the baseline models showed limitations when presented with unseen entities, lengthy inputs, and multiple entities. These challenges highlight the difficulties of dataset mention detection and linking in scientific literature. To develop a more robust mention detection method, future research may also explore end-to-end framework for dataset entity mention detection and linking, advanced detection networks that are robust to noise in training data, or how to leverage the context out of the mention sentence to boost the performance of EL. In addition to these approaches, future work may also explore the use of footnotes and citations in literature to improve dataset entity recognition.

\section{Conclusion}
In conclusion, DMDD is a valuable resource for studying dataset mention detection in scientific literature. As the largest corpus created for this purpose, it addresses the limitations of existing corpora in terms of size, diversity of dataset mentions, and entity linking information. Our experiments with baseline models show that DMDD enables the training of more robust models with a small number of manual labels, as demonstrated by the improved performance of SciBERT trained on DMDD compared to other corpora. The analysis of DMDD instances and experimental results highlight the challenges and open problems in the task of dataset mention detection. We believe that DMDD will stimulate further research in this important area of scientific information extraction.


\section*{Acknowledgement}
This work was supported by the National Science Foundation awards III-2107213 and III-2107518.
We thank the TACL Action Editor and the anonymous reviewers for their valuable comments. We also thank undergraduate students Eric Reizas at Temple University and Kathan Parikh at UIC for their valuable contributions to our project.

\bibliography{main}
\bibliographystyle{acl_natbib}

\onecolumn


\end{document}